# A new TAG Formalism for Tamil and Parser Analytics

**Vijay Krishna Menon**
Centre for Excellence in Computational Engineering and Networking (CEN),
Amrita School of Engineering,
Amrita Vishwa Vidyapeetham, Coimbatore.
M_vijaykrishna@cb.amrita.edu

**Rajendran S**
Centre for Excellence in Computational Engineering and Networking (CEN),
Amrita School of Engineering,
Amrita Vishwa Vidyapeetham, Coimbatore,
rajushush@gmail.com

**Anand Kumar M**
Centre for Excellence in Computational Engineering and Networking (CEN)
Amrita School of Engineering
Amrita Vishwa Vidyapeetham, Coimbatore
m_anandkumar@cb.amrita.edu

**Soman K P**
Centre for Excellence in Computational Engineering and Networking (CEN)
Amrita School of Engineering
Amrita Vishwa Vidyapeetham, Coimbatore
kp_soman@amrita.edu

## Abstract

Tree adjoining grammar (TAG) is specifically suited for morph rich and agglutinated languages like Tamil due to its psycho linguistic features and parse time dependency and morph resolution. Though TAG and LTAG formalisms have been known for about 3 decades, efforts on designing TAG Syntax for Tamil have not been entirely successful due to the complexity of its specification and the rich morphology of Tamil language. In this paper we present a minimalistic TAG for Tamil without much morphological considerations and also introduce a parser implementation with some obvious variations from the XTAG system.

## 1 Overview

TAGs were proposed for language models earlier by Vijay Shankar and Aravind Joshi in (Vijay-Shankar and Joshi, 1985). Unlike the Chomskian formalisms, the elementary objects manipulated by TAG are trees; structured objects and not strings. Such structured formalisms have properties that relate directly to strong generative capacity (structure descriptions), which is linguistically more relevant than string sets (weak generative capacity). So we call TAGs as a tree generating system rather than a string generating system. The set of all trees derived in a TAG constitute the object language. Hence, in order to describe the derivation of a tree in the object language, we will need to know about '*derivation trees*'. The derivation trees are important in both syntactic and semantic senses. TAGs also have some interesting linguistic properties. Lexicalization is one of the key motivations for the study of TAGs, both linguistic and formal. The lexical phenomena now explain many linguistic theories previously thought to be purely syntactic. So the information in lexicons, have increased both in amount and complexity. From the formal perspective, lexicalization allows us to associate every elementary structure (trees) with a lexicon (any word). The famous Greibach Normal Form (also Chomsky Normal Form or CNF) for CFGs is a kind of lexicalization. However it is a weak lexicalization, as the structure of the original grammar is not preserved and all rules cannot be lexicalised. Thus TAGs provide an edge to this errand over conventional CFGs

## 2 Formalism

TAGs were introduced by Joshi et al. (1975) and later Joshi (1985). It is known that tree adjoining languages (TALs) generate some strictly context sensitive languages and fall in the class of the so called 'mildly context sensitive' languages (Joshi et al, 1991). TALs properly contain context-free languages and are properly contained by indexed languages. We will introduce an overview of TAG and then move on to observe the lexicalization process.

A tree-adjoining grammar (TAG), G consists of a quintuple ($\sum$, NT, I, A, S) where

i. $\sum$ is a finite set of terminal symbols. NT is a finite set of non-terminal symbols such that $(\sum \cap NT) = \phi$.

ii. S is a Sentential symbol such that $S \in NT$.

iii. I is a finite set of trees called initial trees, with the following properties
   a. Interior nodes are labelled by non-terminal symbols;
   b. The nodes on the frontier of all initial trees are labelled by terminals or non-terminals; non-terminals symbols on the frontier of any tree in I are marked for substitution which, by convention is a down arrow (↓);

iv. A is a finite set of trees called auxiliary trees, with the following properties
   a. Interior nodes are labelled by non-terminal symbols;
   b. The nodes on the frontier of auxiliary trees are labelled by terminal symbols or non-terminal symbols. Non-terminal symbol on the frontier of trees in A are marked for substitution except for one node, called the foot node; by convention this is marked with an asterisk(*); the label of the foot node must be identical to the root node.

In lexicalised TAG, at least one frontier node must be labelled with a terminal symbol (the anchor) in all initial and auxiliary trees. The set I U A is called the set of elementary trees. If an elementary tree has its root labelled by non-terminal X, then it is called an X-type elementary tree.

A tree built by combining the elementary trees is called derived tree or parse tree. We will now have to understand how the combinations of trees happen as to make a derived tree. There are 2 major composition operations adjoining and substitution.

Adjoining (or adjunction, as it is alternately referred) builds a new tree from an auxiliary tree β and a tree α (α is any tree initial auxiliary or derived). Adjunction has been illustrated in Fig 1. Let 'α' be a tree containing a non-substitution node labeled by X. The resulting tree, γ, obtained by adjoining β to α at node n is structured as:

- The sub-tree of α with root n is displaced by β, along with its root node n.
- The displace sub tree of α will attach itself to β, replacing the foot node of β.

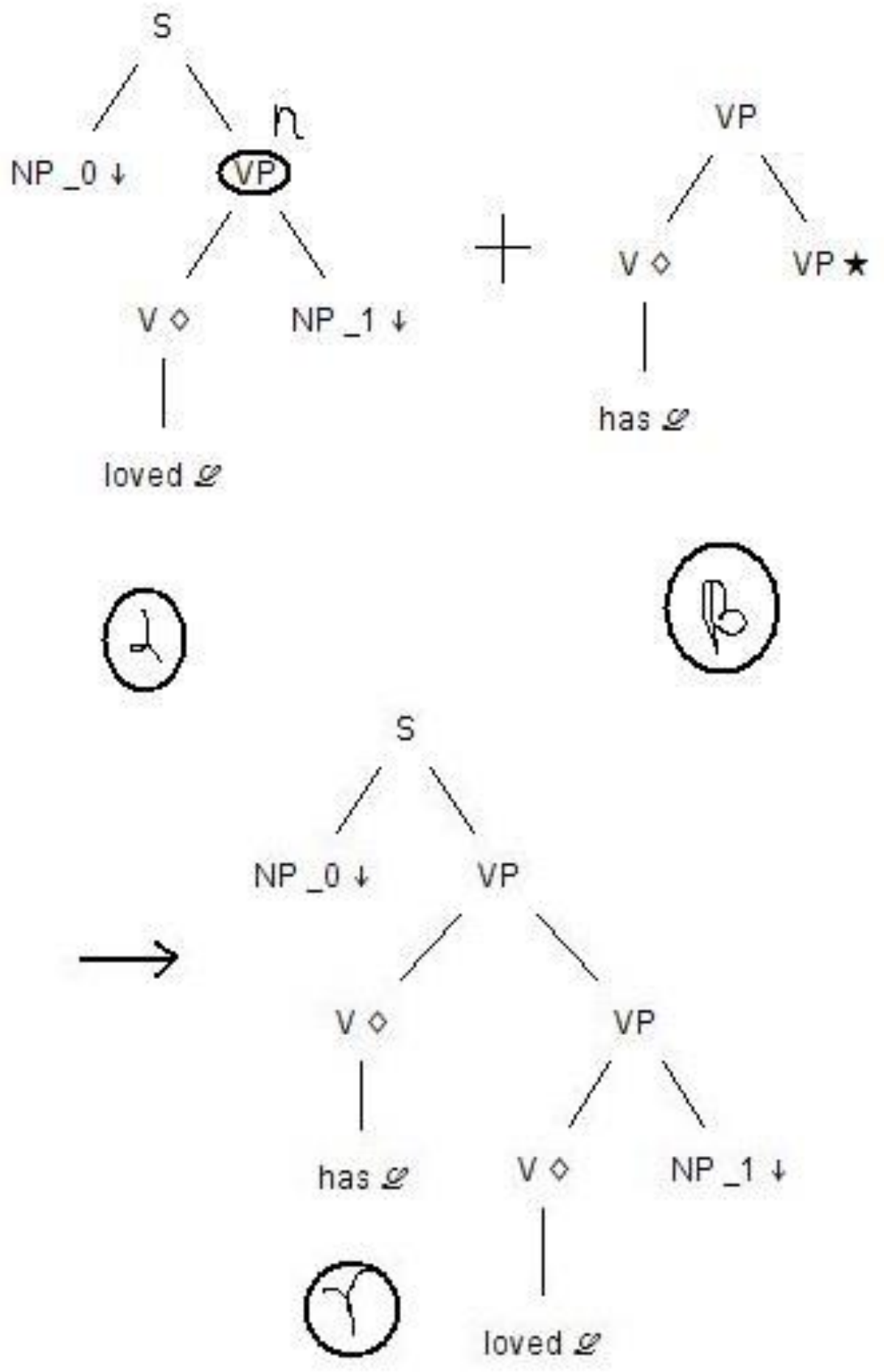

Figure 1: Adjoining of elementary trees

Substitution takes place only on non-terminal nodes in the frontier of a tree. Unlike normal adjunctions, substitutions are mandatory if the node is marked for it with a down arrow as explained above. When a node, say n, is substituted, the entire node is replaced by the initial tree that is substituted. Only initial trees or its derivatives may be used for substitution. By definition adjunctions on any node marked for substitution is not permitted. But adjunctions are possible on the root nodes of the trees already substituted replacing the marked node. This is illustrated in Fig 2 with a set of three initial trees. Substitution extents the targeted leaf node to complete a construct that requires addition of a single substring.

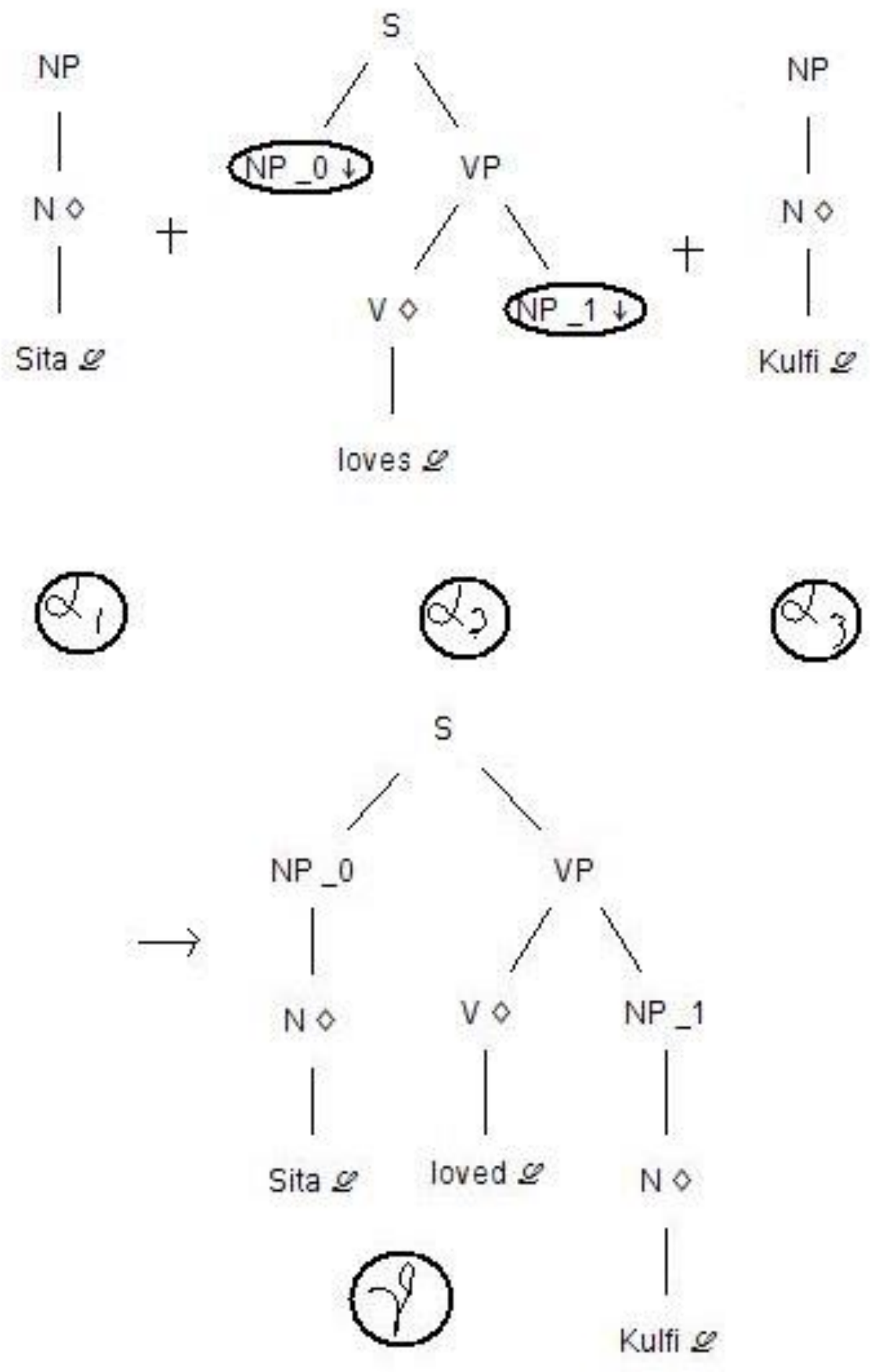

Figure 2: Substituting trees on an initial tree

## 2.1 Adjoining Constraints

Natural language specifics demands more precise ways of adjoining to be used with TAGs. Hence out of the original definition of adjoining, we can add on may constrains that may or may not be applied for an adjunction. Basically an auxiliary tree $\beta$ is adjoined at node n of α, if the root node of $\beta$ is also n (labelled by n) and no substitution is marked on node n of $\alpha$. Now the newer constraints will take effect only on satisfaction of the above basic constraints. They are as follows:

- *Selective Adjunction* (*SA (T),* for short): only elements of the set $T \subseteq A$ of auxiliary trees can be adjoined on the given node. The adjunction is not mandatory.
- *Null Adjunction*[1] (*NA,* for short): It disallows any adjunction on the given node.
- *Obligatory Adjunction* (*OA (T),* for short): the adjunction of any auxiliary tree in $T \subseteq A$ must be mandatorily done on the given node. OA is used to indicate OA (A), which is a common type of Obligatory Adjunction.

If all constraints and substitution operation is withdrawn then the definition of TAG becomes synchronous with the one given in Joshi et al (1975). The latter two additions: constraints and substitution were found to be linguistically useful, hence added with TAGs. Substitution if we see is the main combinatory operation in CFGs. It was introduced by Vijay-Shankar et al (1985). The adjunction constraints allow TAGs to attain some much desired closure properties.

## 2.2 Derivation Structures

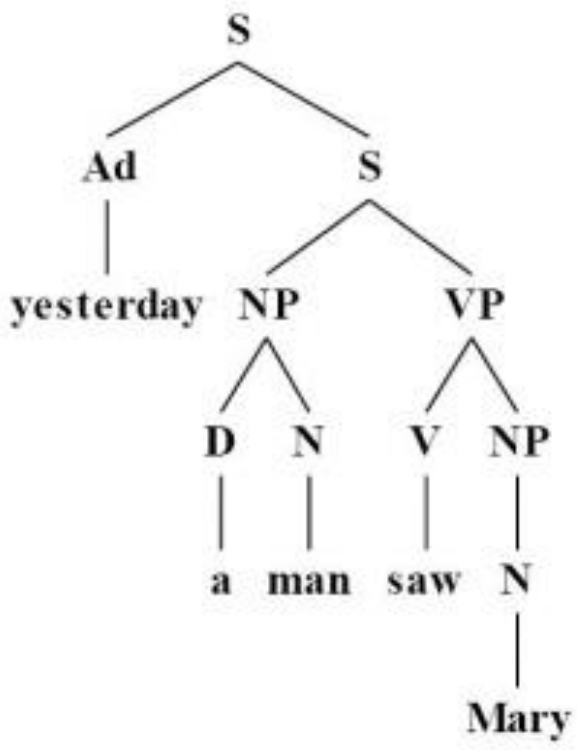

Figure 3

When TAG grammar yields (generates) derived trees by derivation, the information to trace the history of such combination is not given. Unlike CFGs, the derived tree does not contain information as to which basic rules (in our case, elementary trees) were used to construct it. Hence we require a new object that gives us information regarding all operations and elementary trees used to build a derived tree. This structured object is called a *derivation tree*. It uniquely specifies what operation was used to combine which particular trees. Both adjunctions and substitutions are considered for derivation.

Consider the example sentence "*Yesterday a man saw Mary"*. This example has been adopted from Joshi and Schabes (1997). Fig 3 illustrates the derived tree for the above English sentence. But this tree does not give any relevant information regarding how it can be constructed. For this we define the derivation tree for the same sentence. Refer to Fig 4 where the necessary elementary trees required to derive the $\alpha_5$ has been illustrated. Note that α trees are initial trees and the β ones are auxiliary. This convention will be prevailing

[1] Null adjunction corresponds to a special case of selective adjunction SA (T) where T is a null set. i.e., NA ≡ SA(ϕ).

throughout this paper whenever referring to TAG trees.

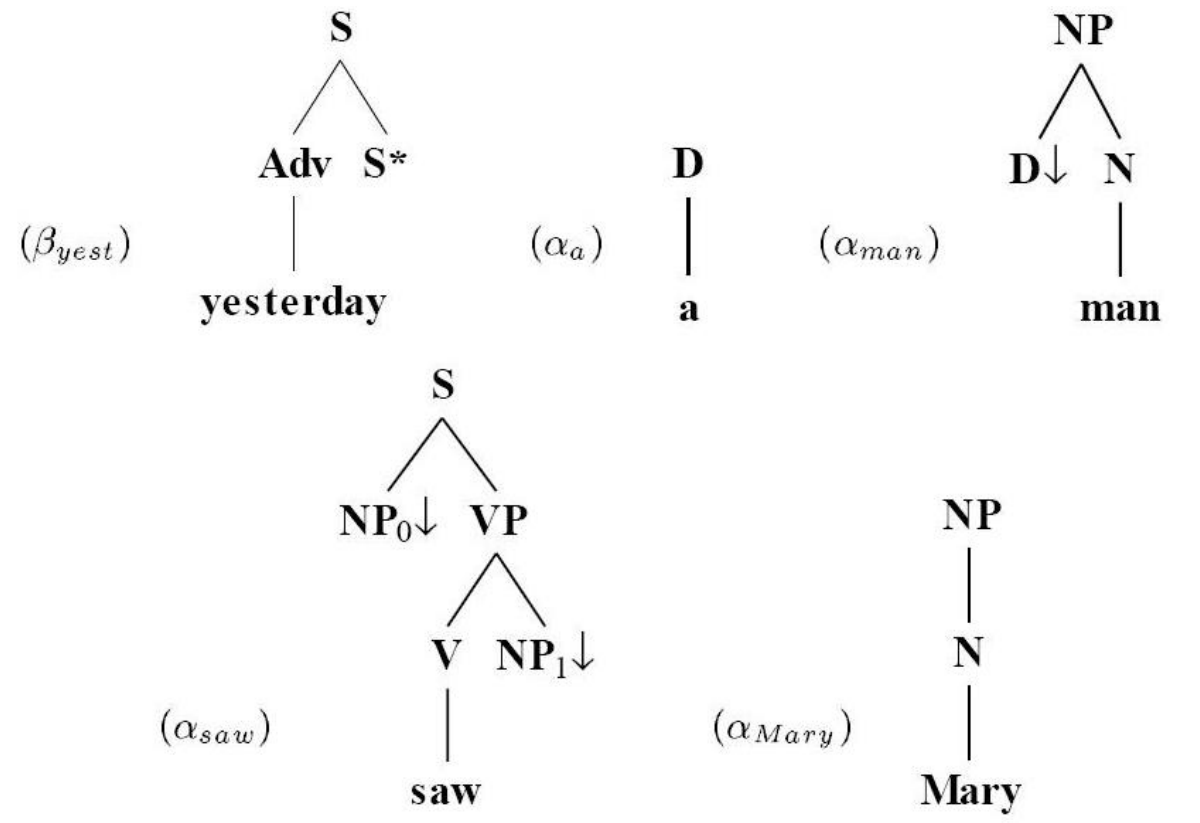

Figure 4: Elementary trees for derived tree in Fig 3

Now the derivation tree for this example is shown in Fig 5. Along with exemplifying the process of building a derivation we also show how a proper lexicalization of TAG is achieved. All the elementary trees in the Fig 4 are properly and completely lexicalised with every elementary tree mapped to at least one lexicon. So every tree will have at least one anchor node.

The roots of all derivation trees are labelled by the name of an S-type initial tree. All child nodes are labelled by auxiliary trees which adjoined or initial trees which are substituted. The notion of tree address is used here to indicate where the composition happened. This will uniquely identify a node in a given tree. This address is referred to as the *Gorn index*; used for multiple array of purposes and is specifically important from an implementation point of view.

The Gorn index system starts with index 0 for the root node. For the 1st level children the numbering starts with 0.1 (or just 1) for the leftmost and increasing towards the right. For the 2nd level children say the child of the second leftmost child will be given 0.2.1 (or just 2.1) and so on. The system is simple and intuitive. Now if an adjunction takes place at this node of the tree, the derivation tree node labelled with the adjoining auxiliary tree will also carry the Gorn index 0.2.1, so we know exactly where the adjunction or substitution has occurred.

Fig 5 depicts the derivation of the derived tree given in Fig 3. Note that $\alpha_{saw}$ is an S-type initial tree; most verb initial trees are expected to be so. Now the node $\alpha_{man}$ *(1)* indicates a substitution of this tree at node *0.1* of $\alpha_{saw}$. In a deeper sense it means this tree replaced the node indexed 0.1 in tree $\alpha_{saw}$.

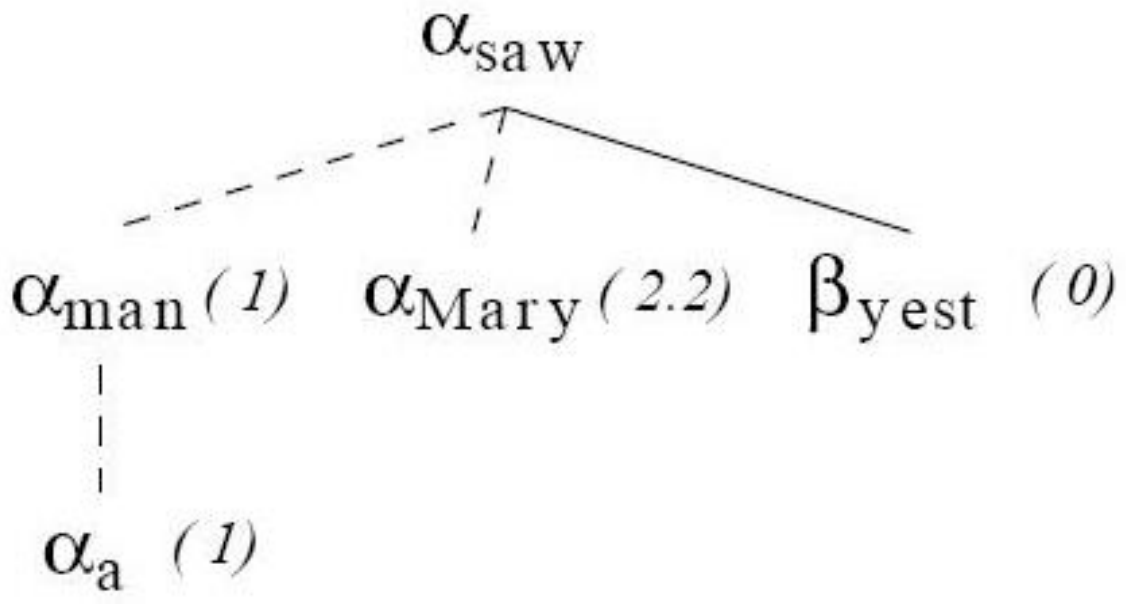

Figure 5: Derivation for the above example

The case with $\alpha_{Mary}$ is no different, except that it is substituted at for node *0.2.2*. But $\beta_{yesterday}$ is an auxiliary tree and is adjoined at the root node of $\alpha_{saw}$ as it contains the Gorn index pointing to the root. The main idea here is the Gorn indices given in a derivation tree's node, points to an address in its parent node's tree where the substitution or adjunction has been done. Further it also demonstrates how lower composition happens, like $\alpha_a$ substituted on $\alpha_{man}$. Unlike as represented, substitutions need not be discriminated with dotted lines alone. The target node tree can solve the conflict by its type as in initial or auxiliary. Another counter intuitive fact is that adjoining happens even at the root node. But controlling adjunctions will help us control the grammars generative ability and restrict the constructs it creates. So every node in the derivation tree will have distinct indices for a given parent node. This way of representing derivation not only captures the syntactic structure of the target tree but also contains semantic dependencies. This has been demonstrated by Joshi and Rambow (1997); they were the first to investigate this property for TAG derivations. Later, Joshi and Rambow (2003) gave a dependency grammar based on TAG formalism. However we shall give a different picture of the same idea here. To illustrate this let us isolate the basic words of the above given example itself. Before we go into detail of this we will need to define dependency functions of each word with respect to the parts of speech (POS) of each word. Consider initially the verb *saw*. Now 'saw' is a *transitive verb*[2], so it will have dependencies in 2 ways, one with its subject and the other

[2] Verbs that require a subject and an object of action are transitive verbs.

with the object. Hence the dependency function will look like this.

$$f[d](saw) = verb[t](sub, obj)$$
$$= verb[t](man, Mary)$$

This show the dependencies of the transitive verb *saw* to depend on the subject as to who or what *saw* to the object as to *saw* whom or what. Logically this function looks like this for *saw*.

$$< Who > \ saw < Whom >$$

This is exactly what we get in the derivation; "*man saw Mary"* giving us the dependency function for *saw* to be *saw (Man, Mary)*. All the other words will have dependencies too as well. As for the Noun *man* the function is different and addresses the number or specificity. That means that nouns have articles or adjectives that describe them. This is their dependency. The above derivation also gives *man(a)* which is the dependency function for the word. **The dependencies of a word can be easily found from the children of the given node in a derivation tree.**

From the above insight, we must gather that *saw* in this example is not just transitive. That is to say it has a subject, an objects and an adverb. Thus the definition of the function should be having an extra parameter, one that specifies time in this case hence we have *saw (Man, Mary, Yesterday)*. This property of TAG derivation greatly helps for representation of agglutinative languages, where the verbal inflection will depend on its subject or object or both. Subject verb agreements are crucial especially in Indian languages.

## 2.3 Lexicalization

According to Joshi and Schabes, (1997), lexicalized grammars are of both linguistic and formal significance. In lexicalised TAG (LTAG), each elementary tree is systematically associated to a lexical item called *anchor*. The grammar should further define a *lexicon*[3] where every lexical item is associated with a finite number of structures, for which that item is the anchor. There are composition operations that describe the building of such structures. Such a grammar is called a lexicalised grammar.

Before we define lexicalization formally we need to refine some thought curves and ideologies referring to anchors, lexical items, lexicons and terminals. We shall give postulates on these, so they are easy to refer back in time when required. Though this postulations deviate from the original TAG definitions given by Vijay-Shankar (1987) and by Joshi and Schabes (1997). This was required for the model in which TAGs are used in Machine Translation and our implementation of it. These deviations do not disturb the formal properties in any manner but gives greater, intuitive, linguistic flexibility at the same time. It also allows for lexicalization easily, without loss of the generality of the normal grammar.

The lexicalization postulates are as follows.

1. The anchor node is a lexical item.
2. The anchor nodes are labelled by special non terminals that we call anchors
3. The anchors represent direct terminal groups, such as POS classes.
4. Adjoining happen normally at anchor nodes unless otherwise specified.
5. All elementary trees carry exactly one anchor for every tree.
6. All trees are thus pseudo lexicalized. Dummy *lexical node* is inserted bellow the anchor node when needed.
7. Anchor nodes are marked with diamond and lexical nodes with a script-L ($\mathcal{L}$). They exist as separate nodal entities.

These postulates will be further deliberated on when we see L-TAG Parsing implementation and analytics. Most of it will only make sense then as keeping these invariants has sped up parsing time complexity. Now we shall proceed to the formal definition of lexicalisation. A grammar is lexicalised if it consists of:

- A finite set of structures each associated with a lexical item; each lexical item will be called the anchor of the corresponding structure;
- An operation or operations for composing structures.

We require the anchor to be a non-empty lexical item. We shall define a *Lexicon*[4] which shall contain or map to a finite set of structures each associ-ated with an anchor, called elementary structures. We will consider operations of combining two structures. These can be restricted as to yield languages of constant growth. The operations that we use will attribute these properties. From the above definition, we see some properties of lexicalised grammars.

[3] Lexicon is an ambiguous name. It will refer to a computer dictionary of all words or it can specifically refer to a single word, a terminal symbol of the grammar. Here it means a dictionary of plausible structures for an anchor.

[4] A lexical dictionary is at times referred to as just the Lexicon. Hence a Lexicon with L capitalized would refer to a lexical dictionary and lexicon with a lower case l would refer to a specific terminal symbol or a word.

***Lemma 1:*** *Lexicalised grammars are finitely ambiguous.*

A grammar is said to be finitely ambiguous if there is no sentence of finite length that can be analysed in an infinite number of ways. This can be seen holding true for lexicalised grammars. Consider any arbitrary sentence of finite length. Since the number words (lexicons) in this sentence are finite so is the number of structures necessary to analyse it. Now a finite number of structures can only be combined in finite number of ways, to produce finitely many structures. Therefore lexicalised grammars are finitely ambiguous.

***Lemma 2:*** *It is decidable whether or not a string is accepted by a lexicalised grammar.*

Lexicalisation does not imply that all grammars be lexicalised. Given any grammar $G$ stated in certain formalism, we can generate $G_L$ though not necessarily in the same formalism, which generates the same tree set, holding the lexical property, henceforth yielding the same string language. This is the lexicalisation phenomenon. From this facsimile the following statement holds true. Joshi and Schabes (1997) says, a formalism $F$ can be lexicalised by another formalism $F'$, if for any finitely ambiguous grammar $G$ in $F$ there is a grammar $G'$ in $F'$ such that $G'$ is lexicalised and generate the same tree set.

***Lemma 3:*** *If G is a finitely ambiguous CFG which does not generate the empty string, then there is a lexicalised tree-adjoining grammar $G_{lex}$ generating the same language and tree set as G such that $G_{lex}$ can have no substitution node in any elementary tree.*

The proof of the above constructivism is given by Joshi and Schabes (1997). It is proved by exemplification where in a grammar is defined satisfying the requirements of this lemma. The main thought for such construction is to separate the recursive part from the non-recursive part of the grammar. For every recursion an auxiliary tree is created; for rule of the type $B \rightarrow \alpha B \beta$ an auxiliary tree will be created.

## 3 TAG Tamil Syntax Mapping

In the tree adjoining frame work, it is understood that each verb in syntactic lexicon selects one tree family. Here we propose to map tree frames for a finite classes of verbs (53 tree families according to one version) in English into Tamil. Tamil being a SOV language is different from English which is an SVO language in the tree configuration. English is both right branching (=head initial) and left-branching (=head-final) language, whereas Tamil is strongly left branching (head final) language. This difference between English and Tamil will be reflected in many structural configurations of English and Tamil. For example, that-complement (that-embedded clause) in English follows the matrix clause, where as in Tamil it normally precedes the matrix clause. Similarly English in the relative clause the head noun comes before the verb, whereas in Tamil the head noun comes after the verb. English is a prepositional language, but Tamil is the post-positional in nature. The post-positions are either suffix or separate particles. When we go by word-alignment, it may appear that the words are just reversed in Tamil when compared to the order of English. We hope to capture these configuration differences between English and Tamil and thereby find ways to map English sentence into Tamil. Here we will map the English tree configuration for a class of verbs with Tamil tree. We hope this will help us when go for transferring English tag configuration into Tamil tag configuration, for the sake of across-the-language NLP applications, for example machine translation. Here will cover up only important tree families into Tamil tree families.

### 3.1 Intransitive: Tnx0V

This tree family is selected by verbs that do not require an object complement of any type. Adverbs, prepositional phrases and other adjuncts may adjoin on, but are not required for the sentences to be grammatical. 1,878 verbs select this family. For example the verbs like eat, sleep, dance, etc., select this tree as illustrated in Fig 6. These **discussions are directly taken from Sarkar (2002)** and not be repeated again and again as reference. Some example sentences: *'Al ate'*, *'Seth slept'* and *'Hyun danced'*. Sarkar (2002)

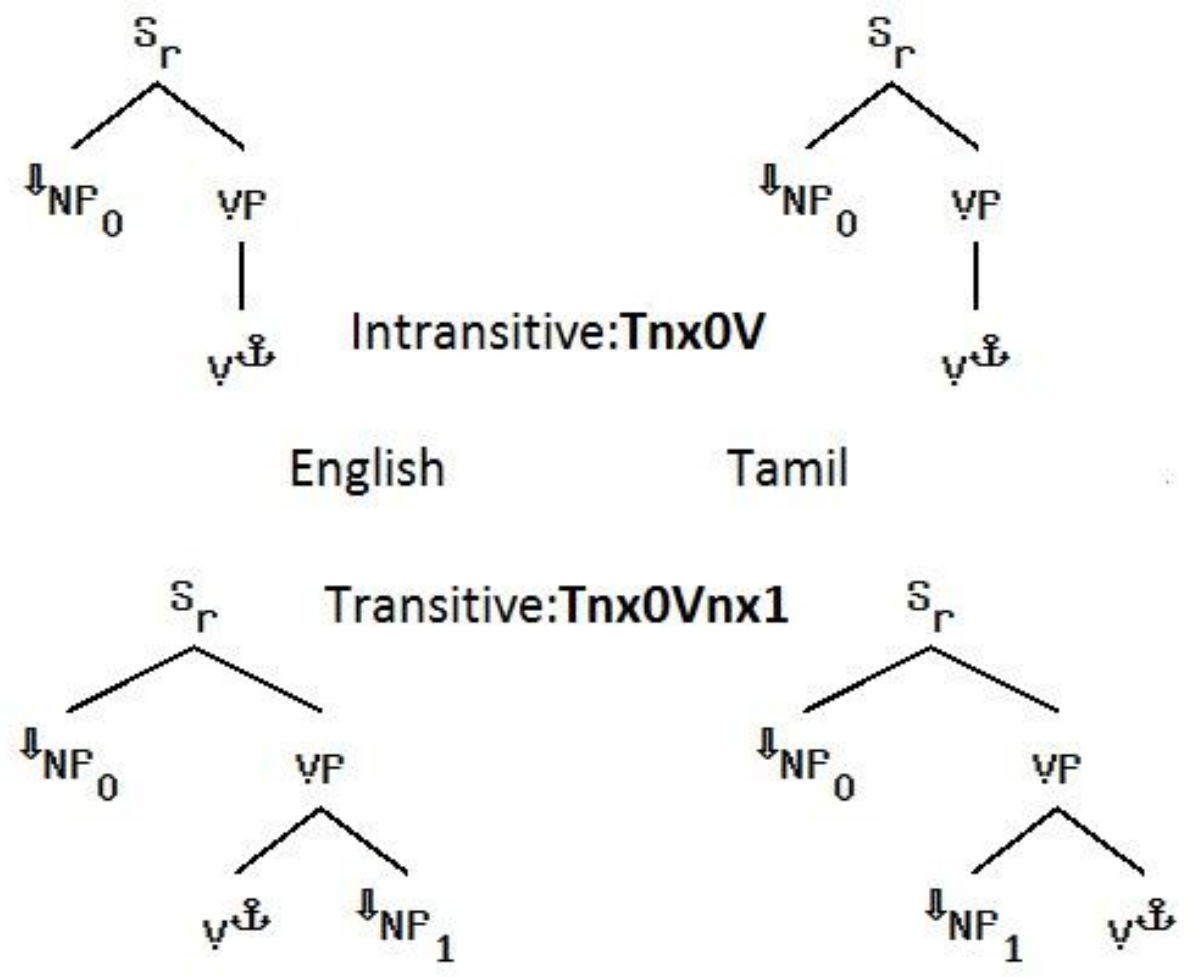

Figure 6: Trees for transitive and intransitive verbs

### 3.2 Transitives: Tnx0Vnx1

This tree family is selected by verbs that require only an NP object complement. The NP's may be complex structures, including gerund NP's and NP's that take sentential complements. This does not include light verb constructions. 4,337 verbs select the transitive tree family. For example the verbs *eat, dance, take, like*, etc. take this tree; sentential examples are: *'Al ate an apple'*, *'Seth danced the tango'*, *'Hyun is taking an algorithms course'* and *'Anoop likes the fact that the semester is finished'*. The Tamil and English mappings are given in Fig 6.

### 3.3 Ditransitive with PP: Tnx0Vnx1pnx2

This tree family is selected by ditransitive verbs that take a noun phrase followed by a prepositional phrase. The preposition is not constrained in the syntactic lexicon. The preposition must be required and not optional - that is, the sentence must be ungrammatical with just the noun phrase (e.g. John put the table). No verbs, therefore, should select both this tree family and the transitive tree family. There are 5 verbs that select this tree family. For example the verbs *ensconce*, *put*, *usher*, etc. select this tree. *Sentential examples: 'Mary ensconced herself on the sofa'*, *'He put the book on the table'* and *'He ushered the patrons into the theater'*.

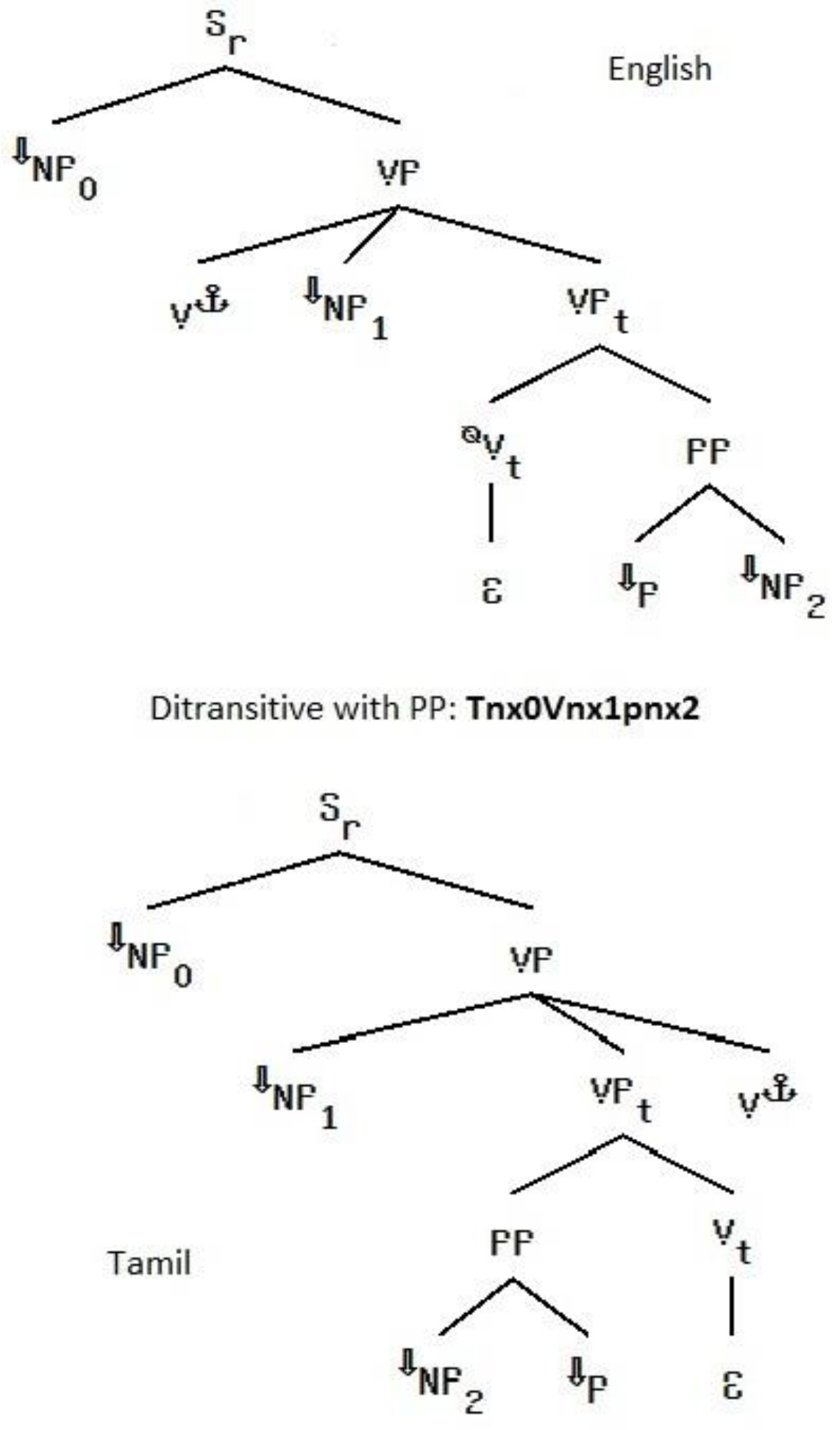

Figure 7: Ditrasitive with PP

## 4 Parsing Analytics and Examples

Vijay-Shankar and Joshi (1985) came up with $O(n^6)$ CYK parser for TAGs. This was the first practical parser for the entire formalism. Later on [Schabes and Joshi, 1988] described their *Earley type TAG parsing algorithm* with extensions to various derived formalisms as well such as constraint TAGs. This was our guiding paper for the modified practical implementation we are to discuss. We have several modifications and deviations that we made to the actual parser. Furthermore, Joshi and Schabes (1997) have proposed a chart parsing algorithm which has been partially adopted by us and in combination came up with a *multi-threaded implementation* of the same for TAGs. The main benefit of an Earley parser is that it has a *worst case complexity* of $O(n^6)$, which in most cases perform much better, as com-pared to the same *average complexity* of the *CYK* parser. However it depends on the grammar definitions, and practically perform better of than in theory. Schabes and Joshi, (1988), starts with the original Earley's invariant and goes on to define one, for the trees, Preserving the actual perspective.

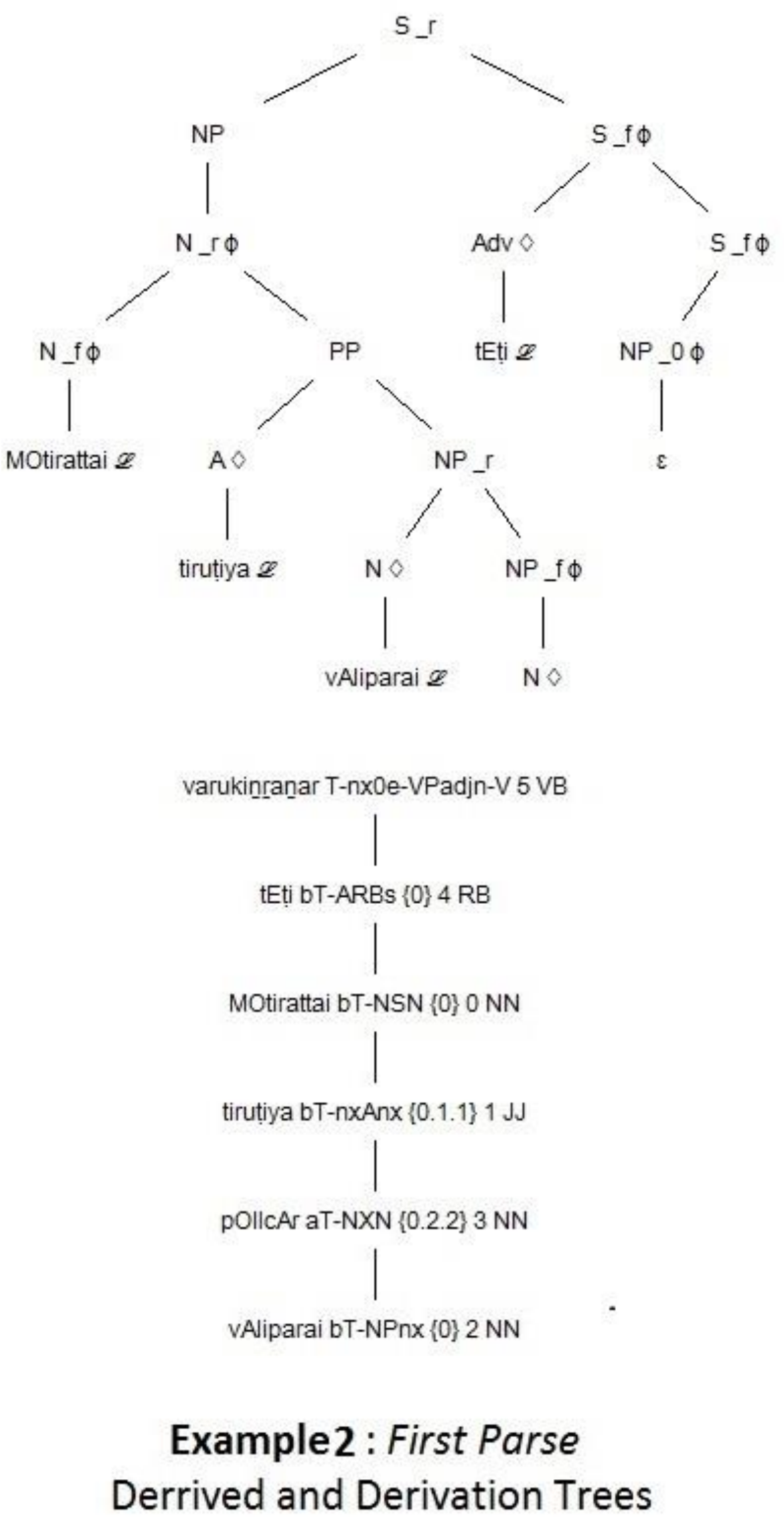

Figure 8: Example 2 First Parse

Since our focus here about the parser is minimal we will not discuss detailed implementation of this parser or the algorithm. We are not using a statistical parser, which is why we will have multiple parses for a single sentence. Mostly this is due to lexicosyntactic ambiguities inherent in the Language specification and the grammar specification. The grammar was made with multilingual tasks in mind mainly machine translation.

## 4.1 Tamil parse specifics and examples

The grammar used by the parser for Tamil contains over 120 trees correctly and is regularly pruned to reduce cross ambiguities. We have 25 initial and 95 auxiliary trees. Together they address most constructs of simple and direct sentences.

The parser accepts Parts of speech tagged sentences using the Penn Tag set for the same. If the sentence is within the construct range of the grammar, the parser immediate returns TAG derivations from which derived trees can be easily constructed. As mentioned before we are not currently dealing with morph analysis just to keep our focus on grammar and parsing. In the examples here the words are mostly surface forms with some chucks in it. Our objective is to prove the conformity of TAG syntax for Tamil in a broader sense. Two examples of the parse instances are illustrated bellow. The Tamil sentences has been Romanised for the sake of linguistic verification.

***Example 1***: NiyUyArkkil naṭaipeRRa yu.Es-OpaN-Aṇkaḷ-iraṭṭaiyar iRutip-pOṭṭiyil liyAṇṭar-payas-jOṭi veRRi peRRu paṭṭattaik kaippaRRiyatu

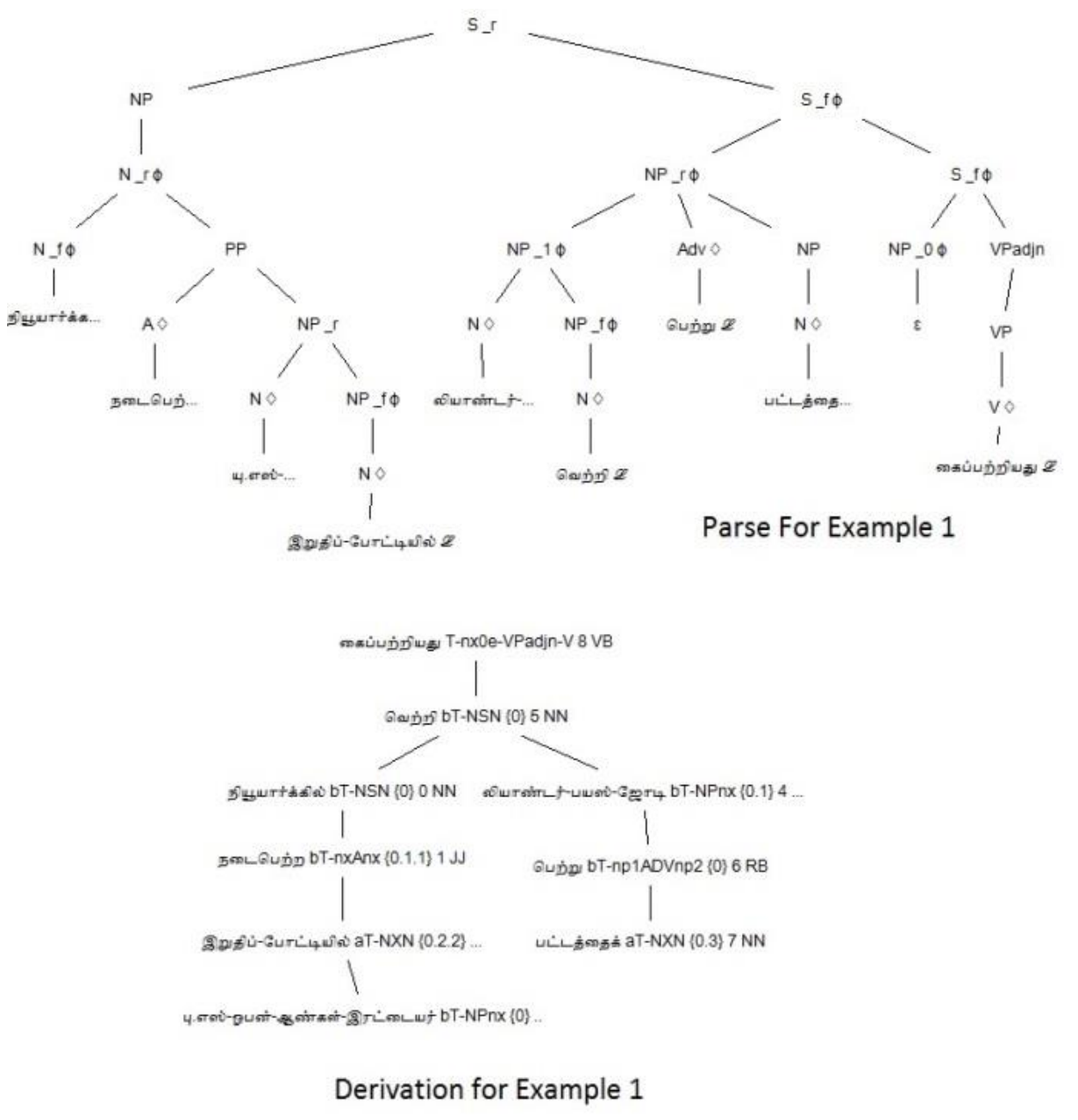

Figure 9: Example 1

***Example 2***: MOtirattai tiruṭiya vAliparai pOlIcAr tEṭi varukiṉṟaṉar

This sentence has multiple parses mainly due to a lexicosyntactic ambiguity. One prominent parse illustrated by Fig 8 and another parse illustrated by Fig 10.

## 5 Conclusion and Further development

Evidently Tamil syntax can be effectively captured using TAG formalism. Even though we have not discussed morphological considerations here we maintain that we can very efficiently deal with it using TAGs. Feature based parsing is very easy as the dependencies are preserved with trees and separate dependency grammars are not required. Our main goal is to apply these techniques in multilingual applications predominantly in machine translation. The synchronised grammar methodology is well suited for such future endeavours.

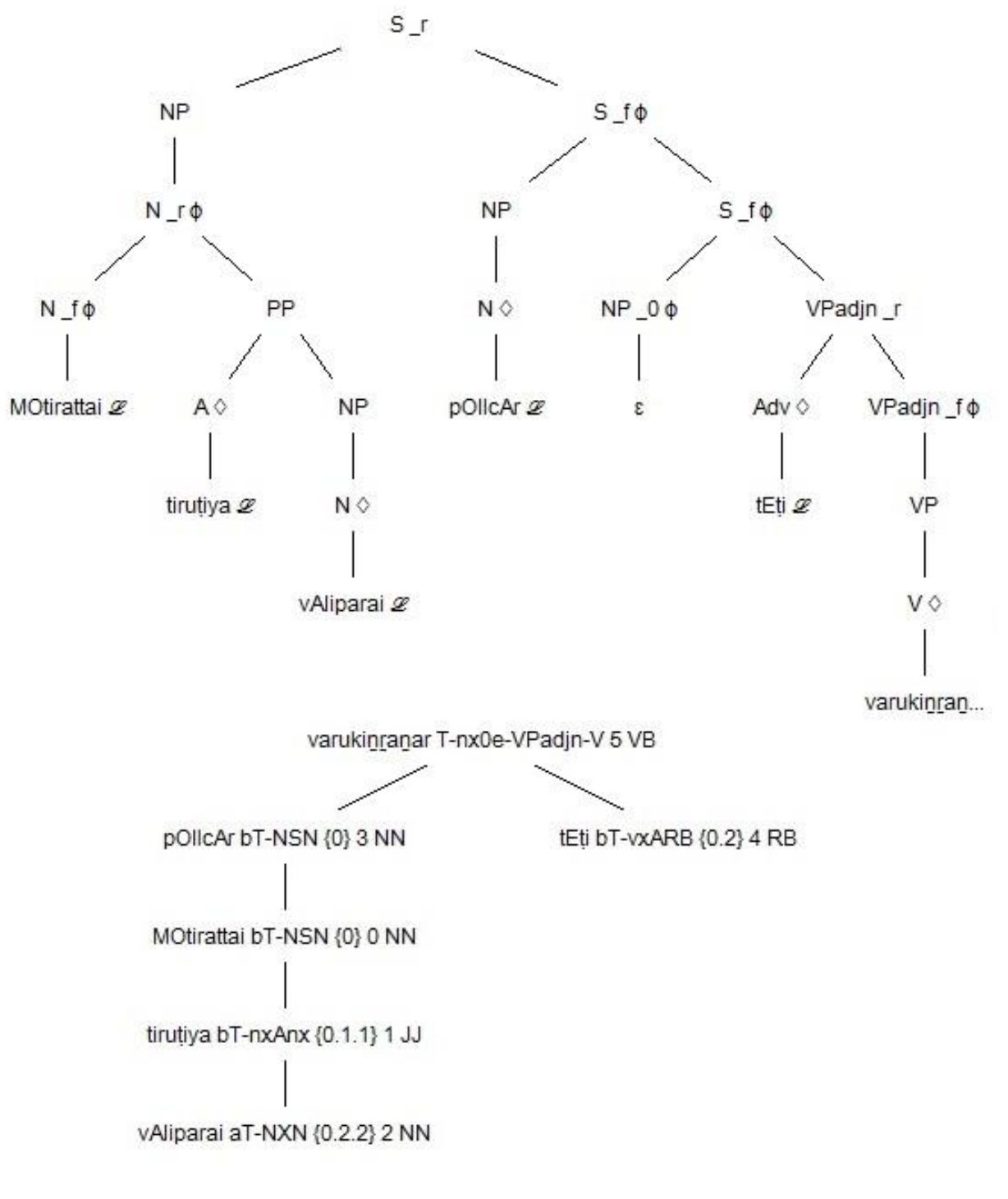

**Example 2:** Sencond Parse
Derived and Derivation Tree

Figure 10: Example 2 Second Parse